\pdfoutput=1
\documentclass[10pt]{cai}

\begin{document}
% Editorial staff will replace the following values:
% 1. Conference Year
% 2. Issue number
% 3. Article DOI
\def\conferenceyear{2024}
\volumeheader{37}{0}%{00.000}
\vspace{-0.6cm}
\begin{center}
\vspace{-0.1cm}

\title{Rumour Evaluation with Very Large Language Models}
\maketitle

\vspace{-0.4cm}
% \thispagestyle{empty}

% Add Authors and Affiliations in the camera ready
% for the double blind review, please leave this section as is 
\begin{tabular}{cc}
% \vspace{-0.1cm}

Dahlia Shehata, Robin Cohen, Charles L. A. Clarke
\\[0.25ex]
{\small  Cheriton School of Computer Science,  University of Waterloo, Canada}
{\small} \\
\end{tabular}

% \vspace{-0.1cm}
% Replace with corresponding author email address
\emails{
  \upstairs{*}dahlia.shehata@uwaterloo.ca 
}
% \vspace*{0.2in}
\end{center}
\vspace{-0.3cm}
\begin{abstract}
% The rapid evolution that generative artificial intelligence (AI) has been witnessing in the last years has impacted a wide range of domains.
% from media and marketing to science and healthcare.
Conversational prompt-engineering-based large language models (LLMs) have enabled targeted control over the output creation, enhancing versatility, adaptability and adhoc retrieval. 
From another perspective, digital misinformation has reached new levels.
% Rumours are thriving in the rapid-fire environment of social media platforms.
% , where unverified information spread instantly and widely.
The anonymity, availability and reach of social media offer fertile ground for rumours to propagate.
% Prior research has mostly relied on the traditional black-box training-finetuning paradigm for rumour classification.
This work proposes to leverage the advancement of prompting-dependent LLMs to combat misinformation by extending the research efforts of the RumourEval task on its Twitter dataset.
To the end, we employ two prompting-based LLM variants (GPT-3.5-turbo and GPT-4) to extend the two RumourEval subtasks: (1) veracity prediction, and (2) stance classification. For veracity prediction, three classifications schemes are experimented per GPT variant. Each scheme is tested in zero-, one- and few-shot settings.
% Two among them did not exist in RumourEval original experiments.
Our best results outperform the precedent ones by a substantial margin. For stance classification, prompting-based-approaches show comparable performance to prior results, with no improvement over finetuning methods. Rumour stance subtask is also extended beyond the original setting to allow multiclass classification.
All of the generated predictions for both subtasks are equipped with confidence scores determining their trustworthiness degree according to the LLM, and post-hoc justifications for explainability and interpretability purposes. Our primary aim is AI for social good.
% we share our code, experimental settings, result analysis, challenges, limitations, and potential future work with the research community.
\end{abstract}

% add your keywords
\begin{keywords}{Keywords:}Misinformation in Social Networks, Large Language Model, Explainable AI
% Generative AI, Explainable AI, Large Language Model, Prompt Engineering, Misinformation, Rumour Evaluation
\end{keywords}
\vspace{-0.3cm}
\copyrightnotice

\vspace{-0.5cm}
\section{Introduction}
\vspace{-0.2cm}

The age of digital connectivity has bridged geographical distances, enabling seamless communication, knowledge sharing across borders and wildfire-like spread of rumours too. 
% Digital misinformation has become a pervasive threat to the modern world.
% Whether on social media platforms or messaging applications, the anonymity of online interactions and the development of algorithms that promote user engagement over content accuracy have amplified the circulation of misleading content with no fear of the consequences.
% Social polarization, institutional distrust, political instability or healthcare fake advises, rumours can have a detrimental impact on all society aspects.
Research efforts have leveraged machine/deep learning (ML/DL) \cite{Patel2022RumourDU, Anggrainingsih2022EvaluatingBP}
% \cite{Hamidian2015RumorDA, Azri2022RumorCT, Pan2022ATR},  \cite {zhou-etal-2019-early, Patel2022RumourDU, Zhao2023PANACEAAA, Anggrainingsih2022EvaluatingBP}
for rumour detection and classification. Without a doubt, DL complex architectures have revolutionized the natural language processing (NLP) research standards for this problem, introducing new baselines. From contextualized pre-trained contextual embeddings \cite{10.1007/s11042-020-10183-2}, transfer learning \cite{9992029}, cross-lingual learning \cite{Tian2021RumourDV}, to multilingual solutions \cite{pranesh-etal-2021-cmta}. Nonetheless, most of these efforts fall under the traditional/predictive AI umbrella that requires long training hours of data.

In another context, generative AI has been evolving tremendously in the last decade.
% Although the concept has been around for years, it was not until recently that its influence spread throughout the globe.
% In healthcare, finance, education, art design, or entertainment \cite{finlayson2019artificial}, generative AI applications are infinite.
This particular subfield of AI is capable of creating original content from scratch, from text to images, and even complex data structures, without relying on similar historical data patterns.
% This advancement was incited by a number of factors, including, but not limited to, the increasing availability of massive amounts of data, the rise of optimized computational algorithms, the rise of cloud computing, the availability of relatively affordable computational resources, and the surging demand for generative AI by diverse industries.
% At its core, generative AI is nothing but sophisticated DL models with cascading multi-layered architectures and billions of parameters that are fed huge amounts of data.
% These models are trained for a pre-defined number of iterations to capture patterns and learn structures.
In this context, the most notable models are
% Generative Adversarial Networks (GANs) \cite{goodfellow2014generative},
Variational Autoencoders (VAEs) \cite{kingma2014auto}, and Transformer \cite{vaswani2017attention} models.
% Introduced in 2017, the Transformer has taken the research community by storm, introducing new state-of-the-art (SOTA) for all the NLP tasks. Transformer-based models are capable of generating text by learning sequential dependencies within data and relying solely on the attention mechanism \cite{vaswani2017attention}.
The advances and optimizations in the Transformers' architecture \cite{vaswani2017attention} have laid the groundwork for the emergence of large language models (LLMs). The latter constitute a pivotal transition in the DL architectural standards. As per the name "LLMs", these models leverage vast amounts of parameters and data to generate more accurate and fluent language representations.
% In the forefront of these LLMs was Google's BERT \cite{bert}, that was the first to learn bidirectional representations of text when the norm was understanding unidirectional representations.
% This breakthrough has opened the door to the development of an infinite list of highly capable LLMs such as Google's BERT \cite{bert, roberta, albert, distilbert} and OpenAI's GPT \cite{gpt2, gpt3-org} families.
In the forefront of these LLMs were Google's BERT \cite{bert} and OpenAI's GPT \cite{gpt2, gpt3-org} families.
% family (i.e. RoBERTa \cite{roberta}, ALBERT \cite{albert}, DistilBERT \cite{distilbert}, XLNet \cite{xlnet}, etc.), OpenAI's GPT family (i.e. GPT-2 \cite{gpt2}, GPT-3 \cite{gpt3-org}, etc.), in addition to other variants like the MEGATRON \cite{shoeybi2019} and T5 \cite{raffel2020} families.
These LLMs are the foundation of the recent advancement of generative AI, but can still be categorized among traditional AI solutions.
As generative models, Very Large Language Models (VLLMs) are the current SOTA for NLP research. The name is attributed to the enormous corpora on which these models were previously trained, in addition to their few trillions of parameters. This latest trend started in 2022 with the emergence of models like Google's PaLM \cite{palm}, OpenAI's ChatGPT
% , InstructGPT \cite{Ouyang2022TrainingLM}
; and 2023 models like Meta's LLaMA \cite{llama} and OpenAI's GPT-4 \cite{gpt4}.

VLLMs are conversational interactive models that rely on prompt engineering to enhance their output and tailor them to a particular context.
% Prompt engineering is the process of crafting the most optimal set of instructions (i.e. prompt) containing task definition, related scope, explicit queries, inputs, expected output format and style, and contextual background to guide the VLLM to the desired responses. It is considered as a bridge between human intentions and VLLM's capabilities.
They allow users to translate their ideas to a format the model can understand, hence the trade-off between specificity and granularity. 
% The more specific a prompt is, the more it can restrict the model's potential to create and explore diverse approaches. The more it is granular, the more the model can deviate from the expected output path, leading to irrelevant or unfocused results.
Although VLLMs have access to a massive external world knowledge in all fields, they can still be customized to a particular context through one-shot or few-shot learning depending on the task complexity \cite{Chen2023UnleashingTP}. One-shot prompting is an approach where the model is given a single example as part of the prompt to help its learning process. Few-shot prompting is when a number of examples are used instead \cite{Chen2023UnleashingTP}. For simpler tasks, zero-shot learning still gives very accurate response, especially with recent VLLMs like GPT-3.5 \cite{gpt3.5} and GPT-4 \cite{gpt4}. It is also worth mentioning that zero-shot setting outperforms one-shot and few-shot in certain occasions. So, the concept of "the more learning examples the better" is a myth \cite{Chen2023UnleashingTP}. Prompt engineering is a process of iterative refinement and feedback.
% In this work, all the experiments are conducted in a zero-shot setting.

Despite the fact that conversational VLLMs are still under development because of problems like hallucinations, biases and privacy concerns, they can still play a crucial role in advancing the field of explainable AI (XAI). The latter aims to uncover the black-box nature of AI model decisions, by making models more transparent, explanatory and trustworthy to humans. In general, XAI methods can be classified into: 1) Ante-hoc, and 2) Post-hoc methods \cite{A2023100230}. The former build the explanations into the AI model itself in the form of representations of the model's internal state. The latter are concerned with providing descriptions of the factors that led to an outcome after decision-making. VLLMs fall under the second category, and they are considered powerful tools for post-hoc personalized explanations.
% For example, ChatGPT can be asked to generate natural language explanations for its outputs or even other AI model outputs. It can also be used for counterfactual explanations by illustrating how the outcome would have changed if the conditions are different.
Additionally, VLLMs can be leveraged to produce confidence scores for classification problems.
% They can easily employ their knowledge of language and context to assess the likelihood that an input belongs to a particular category.
Such information can shed the light on the model certainty and confidence in its own predictions, offering additional explanation and trust into VLLMs decision-making.

Our goal is to leverage the numerous capabilities conversational VLLMs can offer to limit misinformation spread. Towards this objective, we extend the work of RumourEval task that was part of SemEval workshops of 2017 \cite{rumoureval-2017-semeval} and 2019 \cite{rumoureval-2019-semeval}. RumourEval extension is conducted by experimenting variants of generative AI models on a publicly available Twitter-based dataset \cite{dataset} 
% that was originally part of the PHEME project \cite{pheme}
, and previously used in the RumourEval 2017 version for both rumour veracity prediction and rumour stance classification.
% Only the development (i.e. dev) and test subsets are employed in our experiments.
The evaluation adopts the RumourEval 2017's metrics and baselines. To the best of our knowledge, this goal was not previously explored in published literature yet. 
Our contributions can be summarized as follows:
(1) This work is the first to experiment prompt-engineering-based approach for the two subtasks: rumour veracity and stance classification, with two of the strongest VLLM variants nowadays: GPT-3.5-turbo and GPT-4.
(2) Zero-shot experiments are conducted for the 2 subtasks, without requiring long training hours. Crafting the optimal prompts and temperature parameter tuning are done on the dev subset. The classification results are generated on the test subset. One-shot and few-shot settings are also experimented for the rumour veracity subtask.
% using varying number of examples from the training set.
(3) For rumour veracity prediction subtask, three classification approaches (2-way, 3-way and 2-step classifications) are attempted under zero-shot, one-shot and few-shot settings. This is for each VLLM variant with a total of 18 types of experiments for this subtask.
(4) For rumour stance classification subtask, a novel multiclass classification option is introduced allowing  rumour stance to have multiple categories. This multiclass option is additional to the original two schemes of RumourEval: 3-way and 4-way classification. Meaning, a total of six zero-shot experiments is conducted for the two VLLM variants for this subtask too.
(5) We generate a natural language justification for each prediction/classification in the 24 experiments, as XAI post-hoc explanations for human understanding. Such option was not explored in any of the previous RumourEval versions.
(6) We also generate a confidence score for each prediction/classification in the 24 types of conducted experiments to develop an estimate on how trustworthy the provided answer is from the model's perspective. Confidence scores are also produced for each one of the predicted classes of a single rumourous reply in the newly-introduced multiclass stance classification.
(7) Our best results for rumour veracity prediction subtask, achieved by binary classification, outperform the 2017's baseline by more than 25\% for accuracy score, and the 2017's best participating model by 33.2\% for accuracy and 73.5\% for confidence root mean square error (RMSE).
(8) We share our code, in addition to our experimental settings, results, analysis, challenges and potential future work with the research community.

% The rest of the paper is organised as follows. Section \ref{background} reviews RumourEval tasks and related works at the intersection of rumour evaluation and ML/DL/LLMs. Section \ref{methods} describes our choices for the dataset and VLLMs, our approach, and implementation for the two subtasks. The evaluation in Section \ref{evaluation} summarizes the experimental setup, employed metrics, results, findings, and analysis. In Section \ref{discussion},we discuss challenges, limitations, and potential future works before concluding the work in Section \ref{conclusion}.

%%%%%%%%%%%%%%%%%%%%%%%%%%%%%%%%%%%%%%%%%%%%%%%%%%%%%%%%

\vspace{-0.3cm}
\section{Background and Related Works}
\label{background}
% Prior literature can be classified into: (1) RumourEval, and (2) Beyond RumourEval.
% that can be divided itself into (1) published works, and (2) parallel research.

\vspace{-0.1cm}
\subsection{RumourEval}
\label{rumoureval}
RumourEval has been a principal task in a couple of SemEval workshops \cite{rumoureval-2017-semeval, rumoureval-2019-semeval}. The task incorporates two main subtasks. First, rumour veracity prediction is when a rumour is classified to \textit{"true"} or \textit{"false"} with a confidence score between 0-1. Systems were expected to return a confidence score of 0 if a rumour cannot be verified. Two variants were allowed in this subtask: (1) Open variant where additional context such as Wikipedia dumps and URLs was allowed, and (2) Closed variant where only the training and dev sets were permitted. Second, rumour SDQC stance classification subtask focuses on the discourse around a given rumour. The discourse was represented as nested replies where each reply should be categorized to only one of the four categories: (1) support, (2) deny, (3) query, and (4) comment, depending on its stance towards the source rumour, hence the name "SDQC".
The 2017 version \cite{rumoureval-2017-semeval} chose Twitter as its target social media platform. The participating teams employed relatively old ML/DL methods like ensemble learning, Long short-term memory (LSTMs), convolutional neural networks (CNNs) and support vector machines (SVMs). Feature engineering was an essential step to achieve competitive results with the baselines.
On the other hand, the 2019 version \cite{rumoureval-2019-semeval} of RumourEval expanded the Twitter dataset with Reddit posts. Older ML methods had still a fair share like random forests (RF), logistic regression (LR) and SVMs. However, more advanced DL methods marked their presence like sophisticated neural architectures (ex: Bi-LSTMs), the early version of BERT \cite{bert} and pre-trained contextual embeddings using OpenAI's GPT \cite{Radford2018ImprovingLU} and ELMo \cite{peters-etal-2018-deep}, compared to static embeddings
% of Word2vec \cite{Mikolov2013EfficientEO}
in 2017's version.
RumourEval was discontinued after 2019.

\vspace{-0.3cm}
\subsection{Up-to-date Published Research for Social Media Outside of SemEval: 2021 - 2023}
% \vspace{-0.2cm}
% Outside of SemEval, there is an abundance of papers that address misinformation detection, rumour prediction and stance classification.
% we limit our review to papers targeting social media platforms published between 2021 to 2023 to fit the page limit. 

\vspace{-0.2 cm}
\subsubsection{Veracity Prediction}
\vspace{-0.2cm}
In the context of misinformation detection, \citet{10.1145/3442381.3450111} propose a finetuned-based framework to experiment with Transformer-based language models (LMs) on benchmark datasets.
% The pre-trained LMs include models like BERT, EMLo and others.
\citet{kim-yoon-2022-detecting} 
% introduce an interesting definition for rumours being an "
% information that cannot be verified at the time
% of origination", and
propose a BERT-based double-channel structure to determine the ex-ante veracity of rumours on social media.
% The first channel categorizes the rumour based on its certainty, then the certain ones are evaluated by a BERT-based lie detector. 
\citet{Jain2022CanarDeepAH} propose CanarDeep model for rumour detection, a hybrid DL model that combines the predictions of a hierarchical attention network (HAN) and a multi-layer perceptron (MLP).
% learned using context-based and user-based features.
Experiments are conducted on the benchmark PHEME dataset.
% \citet{Whitehouse2022EvaluationOF} examine the integration of existing knowledge bases (KBs), like Wikidata, into pre-trained LMs.
For multilingual models, \citet{Tian2021RumourDV} present a zero-shot cross-lingual transfer learning framework based on multilingual BERT.
% without the need for labeled data.
% The proposed framework is able to adapt a rumour detection model trained for a source language to another target language.
% \citet{10021113} suggest a Multilingual Source Co-Attention Transformer (MUSCAT), that builds on a multilingual pre-trained LM to perform multilingual rumour detection.
For more customized social media datasets, \citet{KOCHKINA2023103116} create a new COVID-Rumour Verification (COVID-RV) dataset from online tweets.
% , and evaluate the top-performing (at that time) rumour verification pre-trained LMs on the dataset.
Recently, \citet{Pelrine2023TowardsRM} attempt GPT-4 for veracity prediction in zero-shot experiments. While this work may coincide with ours, our experiments include both GPT-3.5 and GPT-4 in three different settings zero-shot, one-shot and few-shot, and for both RumourEval subtasks.

\vspace{-0.4cm}
\subsubsection{SDQC Stance Classification}
\vspace{-0.2cm}
% \citet{stance1} introduce a hybrid of capsule neural network and multi-layer perceptron. 
\citet{doi:10.1177/0165551520944352} propose a framework for identification of rumour stances, combining network topology and social media comments. 
% They built a vector matrix of comments and words via term frequency–inverse document frequency (TF-IDF) optimisation. Another vector matrix is constructed to represent network topology among the users.
A VAE is then applied for dimensionality reduction before matrix-matrix integration.
\citet{adaptivezahra} attempt to find solutions to the well-known "imbalanced class" problem of the stance classification, by proposing a novel adaptive cost-sensitive loss function for learning imbalanced data using DL.
\citet{ScartonL21} introduce the first study
of cross-lingual rumour stance classification through experiments with zero-shot and few-shot
learning and in three languages (German, Danish and Russian). 
% The authors leverage multilingual BERT and machine translation (MT) for rumour analysis in languages where labeled data is not available.
The work of \citet{Li2023EvaluatingTR} proposes using adversarial attacks in test data
to consider that a source tweet may not be as important as expected considering the stance of its related discourse.
% revisits an interesting idea suggesting that the source tweet may not be as important as expected when determining the stance of its related discourse. The authors propose adversarial attacks in the test data.
% to check the veracity of this idea.
% in the social media context.

\vspace{-0.3cm}
\subsubsection{Both Subtasks}
\vspace{-0.2cm}
Among the rarest works that address the two subtasks,
\citet{10.1145/3430984.3431007} proposes a
multi-task learning framework by finetuning Longformer \cite{longformer} model for jointly predicting rumour stance
and veracity on the RumourEval 2019's dataset.
As one can see, all of these research works follow the finetuning paradigm using pre-trained LLMs. VLLMs like OpenAI's GPT-3.5/GPT4, Google's PaLM/Bard and Meta's LLaMA are rarely explored in the published research. Outside SemEval, most of prior works addressed one of the two subtasks only.
% except \cite{10.1145/3430984.3431007}. 

% \vspace{-0.2cm}
% \subsection{Parallel Research: Unpublished yet}
%  Our literature review is by no means exhaustive. However, to the best of  our knowledge, there is no published work leveraging generative AI, conversational VLLMs or prompt-engineering in general, for rumour veracity prediction or stance classification. Nonetheless, we found a parallel research line that is still unpublished attempting GPT-4 on the first subtask only: veracity prediction, in the paper of \citet{Pelrine2023TowardsRM}.

%%%%%%%%%%%%%%%%%%%%%%%%%%%%%%%%%%%%%%%%%%%%%%%%%%%%%%%%

\vspace{-0.3cm}
\section{Methods}
\label{methods}
\vspace{-0.2cm}
Our methods encompass the VLLM choice, dataset choice, design and implementation details for the 2 subtasks in terms of approach, prompt crafting and prediction function.

\vspace{-0.3cm}
\subsection{VLLM Choice}
\vspace{-0.1cm}

Two types of VLLM variants are exploited in this work: GPT-4 \cite{gpt4} and GPT-3.5-turbo \cite{gpt3.5}. Besides the following reasons explaining why we chose each model, it is worth mentioning that "availability" is one of the important factors. OpenAI models were always available in Canada, unlike Google's Bard, for example, that became available very recently. \\
(1) \textbf{GPT-4} is the most capable GPT model series to date, and is the strongest available for many NLP tasks already. The model is able to perform complex tasks, but is generally slower at giving answers and more expensive with known issues of hanging in the middle of a task. It can handle more tokens than its predecessors and can analyze images. GPT-4 is adopted by ChatGPT Plus. \\
(2) \textbf{GPT-3.5-turbo} is the best model in the GPT-3.5 series. It is currently employed by the free version of ChatGPT. Compared to GPT-4, it is cost-effective, fast and flexible. Although it cannot handle images, its generated answers for natural language tasks are on the same par with GPT-4 with slight performance degradation.\\
In addition to their availability, GPT models have well-known tunable parameters (unlike other models) that can be used to improve the model output.
One of those parameters is the "temperature", with a value between 0 and 1, that serves as a control mechanism. The lower the temperature the more consistent the answers are, which makes them reliable and determined with no fluctuations. The higher the temperature, the more random and creative GPT can get when generating its answers.

\vspace{-0.3cm}
\subsection{Dataset Choice}
\vspace{-0.1cm}
We employ the same dataset \cite{dataset} of SemEval 2017: RumourEval \cite{rumoureval-2017-semeval} for the two subtasks. The main reason is to facilitate the comparison with the RumourEval 2017's participating models and baselines, especially with the faced challenges described in \ref{challenges}. In addition, it is publicly available.
% \footnote{https://alt.qcri.org/semeval2017/task8/index.php?id=data-and-tools}.
The authors even share the scoring scripts and the gold standard test data for the 2 subtasks. 
The data had been previously annotated for veracity and rumour stance following the PHEME project \cite{pheme}. The ground truth of the manually labelled tweets by human annotators is used to evaluate the model output to a given prompt.
The original dataset includes 3 components described below:
(1) The training-dev combined dataset encompasses English rumourous threads collected from Twitter for 8 well-known events
% like the Charlie Hebdo shooting in Paris, the Ferguson unrest in the US, and the Germanwings plane crash in the French Alps
\cite{rumoureval-2017-semeval}. 
 % Our experiments in this work are zero-shot, meaning that the training data is of no importance. However, the dev subset is employed for tuning the VLLM temperature and crafting the optimal prompts.
(2) The test set includes 28 additional threads: 20 from the same events of the training set, and 8 from two new events described in \cite{rumoureval-2017-semeval}. The size of the test data is 1080 tweets with 28 source tweets and 1052 replies. All of our results are reported on the test set.
(3) There is an additional context data that participants could leverage incorporating Wikipedia articles and content of linked URLs. This additional context was allowed to be used for the Open variant of the rumour veracity task that was described in \ref{rumoureval}. In  our experiments, we have not used any additional data. This choice was partially because we wanted to experiment VLLM capabilities with no extra help while relying on its internal knowledge only. It would be interesting, though, to enrich the prompt with such info, and compare the performances.

\vspace{-0.3cm}
\subsection{Subtask1: Rumour Veracity Prediction}
% The goal of this subtask is to predict the truthfulness of a given tweet rumour.
\vspace{-0.2cm}
\subsubsection{Approach}
\vspace{-0.1cm}
The goal is to predict the truthfulness of a given tweet rumour. The train and dev sets are first pre-processed to combine all the sparse information related to one data example in a single table row. That is done for every data point to facilitate information drop-in in the prompt later as thoroughly described in \ref{subtask1_prompt}. Listing \ref{lst1} shows one pre-processed example from the dev subset after combining all the related info: journal, id, text and class (i.e. ground truth) in a single row. The same pre-processing is also conducted on the test subset.
\begin{lstlisting} [label=lst1,  language=HTML, caption= Subtask 1: Data Pre-processing.]
{ 'Journal': 'ebola-essien',
  'tweet_id': '521346721226711040',
  'tweet_text': 'Unconfirmed reports claim that Michael Essien has contracted   
                 Ebola. http://t.co/VASQrZdLhH',
  'tweet_class': 'false'}
\end{lstlisting}
\vspace{-0.2cm}
After data pre-processing, 3 types of experiments are conducted: zero-shot, one-shot and few-shot.
Few examples from the pre-processed train set are leveraged in the few-shot and one-shot experiments in an attempt to teach the VLLM by example.
The dev set is used to tune the temperature and craft the optimized prompts as described in \ref{subtask1_prompt}. 
Then, the crafted prompt with the tuned temperature is used to perform classification on the test set. We experiment with 3 types of classification schemes: \\
(1) \textbf{2-way/Binary:} where a rumour is classified to \textit{"true"} or \textit{"false"}. \\
(2) \textbf{3-way/Ternary:} where a rumour is classified as \textit{"true"}, \textit{"false"} or \textit{"unverified"}. \\
(3) \textbf{2-step:} is new to RumourEval 2017 setting. The first step categorizes the tweet as \textit{"verified"} or \textit{"unverified"}. Then, the verified tweets are classified to \textit{"true"} or \textit{"false"}. \\
For each of prediction, the prompts request two additional outputs: \\
(1) \textbf{Confidence Score:} a value between 0 and 1 that measures the certainty degree the model has in its generated answer. Such information offers many advantages such as filtering out low-confidence predictions, and providing trust and explainability to human users. Confidences scores are also generated to keep consistency with prior finetuned-based models.\\
(2) \textbf{Natural Language Justification:} that acts as post-hoc explanations to users containing why the model has chosen a particular class. This information is novel to RumourEval and is suggested in an attempt to allow XAI for this task.\\
As previously described in \ref{rumoureval}, this subtask had originally 2 variants: \textit{Open} and \textit{Closed}. While no additional contextual information was used in our methods, it is not possible to consider these types of experiments \textit{'Closed'} since GPT VLLMs are already trained on massive amounts of data in all domains, so there is always this possibility that their external knowledge may coincide with the additional context data, even without explicitly trying it.

\vspace{-0.4cm}
\subsubsection{Prompt Crafting}
\label{subtask1_prompt}
\vspace{-0.2cm}

The prompt is the set of instructions used by the VLLM to output the desired response. Prompt refinement and improvement was conducted through iterative attempts of trial and error on the dev set. Listing \ref{lst2} shows an example prompt for 3-way classification in a one-shot setting (i.e. using a single training example). The parameterized prompt defines the task and summarizes the instructions that the VLLM needs to follow to be successful. Each classification method (binary, ternary, 2-step) has its own prompts per experiment type (zero-, one- and few-shot). Similarly, each classification scheme has its corresponding prediction function that describes the main operation and lists the number, format and type of the expected outputs. A prediction function requests 3 main outputs: (1) a prediction which is an array of strings that can take values from {\textit{"true"}, \textit{"false"}, \textit{"verified"}, \textit{"unverified"} according to the required classification type, (2) a confidence score which is a number, and (3) a justification which is a text string.
After defining each prompt and predictive function for a given experiment, they are fed to OpenAI's GPT API Call along with inputs tweets.
\begin{lstlisting} [label=lst2, language=HTML, caption={Subtask 1: Prompt Crafting using 3-way Classification in a One-shot Setting.}]
###Rumour Veracity Prediction###\n
###Instructions###\n
Categorise the following tweet text between quotations "{}" to one category 
from [true, false, unverified].\n
Below is a description of each class: \n
true: if the tweet is verified to be true and factual from reliable
      crowdsourced data sources, even beyond the tweet text. \n
false: if the tweet is verified to be a false rumour or a misinformation,
       from any possible data sources. \n
unverified: if the tweet cannot be verified from data sources.\n
Give a confidence score between 0 and 1 for the predicted value. \n
And justify the prediction choice.
Below is an example of a tweet and its corresponding category prediction to
learn from it: \n
tweet: Unformed Russian Embassy staff in London have left for Russia Rumours
       Putin HAS DIED! http://t.co/zSIV8w6FJ2 via @ShaunyNews #PutinDead?
prediction:  false
\end{lstlisting}

\vspace{-0.5cm}
\subsection{Subtask2: Rumour SDQC Stance Classification}
The goal of this subtask is to label the reply tweet according to its interaction with the source tweet (classifying as \textit{"support"}, \text{"deny}, \textit{"query"} or \textit{"comment}).
Our experiments for stance classification are zero-shot only.
% The goal of this subtask is to label the reply tweet according to its interaction with the source tweet. As preciously explained in \ref{rumoureval}, the rumour stance can be classified into one of 4 categories only: support, deny, query and comment. Our experiments for SDQC stance classification are zero-shot only.
\vspace{-0.3cm}
\subsubsection{Approach}
\vspace{-0.2cm}
Similar to the veracity prediction subtask, the dev set is first pre-processed to combine all the information related to a particular data example in a single row. The main difference from subtask 1 is that information for both the source tweet and the related replies are retained.
Listing \ref{lst5} shows a pre-processed example from the dev subset after combining all the info for one data point: journal, tweet id, tweet text and tweet class, reply id, reply text and reply class into a single row. The word "class" here refers to the ground truth (i.e. the human annotation).
The same pre-processing is also conducted on the test subset.
\begin{lstlisting} [label=lst5, language=HTML, caption={Subtask 2: Data Pre-processing.}]
{ 'Journal': 'putinmissing',
  'tweet_id': '576323086888361984',
  'tweet_text': 'This appears to be the original source of the Putin/Kabayeva
                /Swiss clinic rumour http://t.co/bKfJTSGdca',
  'tweet_class': 'unverified',
  'reply_id': '576327374628917248',
  'reply_text': '@BBCDanielS @EllenBarryNYT Well, Blick is a tabloid, so
                 I wouldn't necessarily believe everything they say :-)',
  'reply_class': 'support'}
\end{lstlisting}
\vspace{-0.2cm}
Examples from the pre-processed dev set are leveraged to tune the temperature and optimize the prompts as described in \ref{subtask2_prompt}.
Then, the crafted prompt performs the classification on the test set. We experiment with 3 types of classification schemes: \\
(1) \textbf{3-way:} where the stance of a tweet reply is classified as \textit{"support"}, \textit{"deny"} or \textit{"query"} only, excluding \textit{"comment"}. From RumourEval 2017, it is known that the dataset is skewed towards comments. 3-way classification method was previously employed to prevent overfitting problems in the training/finetuning paradigms.
Although the generative VLLM mechanism is quite different, we thought we could still replicate this approach using VLLMs.\\
(2) \textbf{4-way:} a tweet reply stance is classified as \textit{"support"}, \textit{"deny"}, \textit{"query"} or \textit{"comment"}. \\
(3) \textbf{Multiclass:} Every example can be assigned one or more category among SDQC. One of the shortcomings we found in the prior RumourEval tasks is that stance classification was matched to only one category. By inspecting the tweets and corresponding replies, there are cases where it is hard to pick one side or another. This is also reflected by the inter-annotator agreement for the labeled ground truth in this subtask where it reaches 62.2\% compared to 81.1\% for subtask 1 \cite{rumoureval-2017-semeval}. \\
The prompts request two additional outputs for each generated prediction:\\
(1) \textbf{Confidence Score:} a value between 0 and 1 that measures the certainty and confidence the model has in its answer, hence providing trust and explainability to human users. It is worth mentioning that for the case of multiclass classification, a array of confidence scores is generated reflecting all the predicted classes for each tweet reply. \\
(2) \textbf{Text Justification:} as post-hoc explanations of XAI.

\vspace{-0.3cm}
\subsubsection{Prompt Crafting}
\label{subtask2_prompt}
\vspace{-0.2cm}
Examples from the dev set are used to refine the prompt in an iterative process. Listing \ref{lst7} shows an example prompt for rumour stance multiclass classification. The parameterized prompt defines the task and summarizes the instructions that the VLLM needs to follow. Class definitions for [\textit{"support"}, \textit{"deny"}, \textit{"query"}, \textit{"comment"}] are consistent with the ones of RumourEval 2017 \cite{rumoureval-2017-semeval}. Both the reply tweet text (that needs to be categorized) and the corresponding source tweet are fed to the parameterized prompt.
Each classification method has its corresponding prediction function that describes the number, format and type of the expected outputs. The function requests 3 outputs: (1) a prediction array having SDQC class values, (2) confidence scores which is also an array of numbers corresponding to the predictions (such as in the case of multiclass classification), and (3) a text justification.
\begin{lstlisting} [label=lst7,, language=HTML, caption={ Subtask 2: Prompt Crafting for Zero-shot Multiclass Classification.} ]
###Rumour SDQC Stance Classification has the following 
stance categories: ###\n
support: the author of the response supports the veracity of the rumour
         represented in the source tweet.\n
deny: the author of the response denies the veracity of the rumour
      represented in the source tweet.\n
query: the author of the response asks for additional evidence in relation to
       the veracity of the rumour.\n
comment: the author of the response makes their own comment without a clear
         contribution to assessing the veracity of the rumour.\n\n
###Instructions###\n
Categorise the following tweet reply "{}" \n
to one or more stance categories based on the following
source tweet "{}". \n
Give a confidence score between 0 and 1 for the predicted 
value. \n And justify the prediction choice.
\end{lstlisting}

% \subsubsection{Prediction Function}
% Each classification method has its corresponding prediction function that describes the number, format and type of the expected outputs. The function requests 3 outputs: (1) a prediction: which is an array of strings that can take True/False values, (2) confidence scores which is also an array of numbers corresponding to the prediction array in the case of multiclass classification, and (3) a text justification which is a text string.

% The last step is to pass the input tweets, parameterized content of Listings \ref{lst6} or \ref{lst7} and the prediction function to OpenAI's GPT API Call.

%%%%%%%%%%%%%%%%%%%%%%%%%%%%%%%%%%%%%%%%%%%%%%%%%%%%%%%%

\vspace{-0.6cm}
\section{Evaluation}
\label{evaluation}
% The evaluation addresses the experimental setup, metrics, results and analysis. 
\vspace{-0.2cm}
\subsection{Experimental Setup and Settings}
\vspace{-0.1cm}
Overall, 24 types of experiments (18 for veracity prediction and 6 for SDQC classification) are conducted using GPT-3.5-turbo and GPT-4. A temperature value of 0.8 is found to achieve the best results after tuning on the dev set. For veracity prediction, we attempt zero-, one-, and few-shot experiments. For the latter, a varied number of examples (2, 4, 6) was experimented before concluding that the fewer number of guiding examples, the better. We only report our few-shot best results.
Concerning the SDQC task, only zero-shot experiments were conducted because of our results, findings and analysis of the prior task.
Experiments are conducted on Google Colab without the need of additional CPU or GPU resources.
The implementation is in Python following OpenAI settings and guidelines.
% The experiments are conducted on Google Colab
% \footnote{https://colab.google/}
% without the need of additional CPU or GPU resources. The implementation is written in Python following OpenAI settings and guidelines.
% \footnote{https://platform.openai.com/docs/guides/legacy-fine-tuning/specific-guidelines}.
We also publish our code\footnote{\url{https://github.com/Dahlia-Chehata/RumourEval-with-VLLMs}}
as open-source to the research community.

\vspace{-0.3cm}
\subsection{Metrics and Baselines}
\label{metrics}
\vspace{-0.2cm}
We adopt the same metrics of RumourEval 2017 \cite{rumoureval-2017-semeval} for consistency and easiness. For subtask 1 of veracity prediction, the macroaveraged accuracy is used to describe the ratio of instances having correct predictions. Confidence Root mean square error (RMSE or $\rho$) denotes the difference between system and reference confidence in the correct examples.
Incorrect examples have an RMSE of 1. RMSE ($\rho$) is then normalised and combined with the macroaveraged accuracy to give a final score: $acc = (1 - \rho) * acc$.
We also utilize the same baselines of 2017 experiments, in addition to the open-source scoring scripts\footnote{\url{https://alt.qcri.org/semeval2017/task8/index.php?id=data-and-tools}}.
The baseline is the most common class for subtask 1, and classification accuracy for subtask 2.
In addition to automated scoring, we also do manual inspection on few examples (as shown in Listing \ref{lst12}), especially for new classification approaches, like multiclass classification.

% In addition to automated scoring, we also do manual inspection on few examples (as shown in Listings \ref{lst9} to \ref{lst12}) for each classification type. This manual examination can be effective, especially for new classification approaches that were not previously experimented and do not have a ground reference for comparison, like the case of multiclass classification.

%%%%%%%%%%%%%%%%%%%%%%%%%%%%%%%%%%%%%%%%%%%%%%%%%%%%%%%%

%------------------------------------------------------
\vspace{-0.3cm}
\subsection{Results and Analysis}
\label{results}
\vspace{-0.2cm}

\begin{table}[!t] 
% \vspace{-1cm}
\Small
\caption{Rumour Veracity Prediction Results: The first 6 rows are the participating teams in RumourEval 2017. The baseline shows the upper bound announced in 2017. Our best Score and Confidence RMSE numbers are accompanied with an asterisk.}
\vspace{-2em}
% \centering      % used for centering table 
\begin{tabular}{cc|cc|cc} 
% \hline          %inserts double horizontal lines
% \cline{2-9}
 \\ [0.3ex]
\hline
\textbf{Variant}&\textbf{Model}&\textbf{Classification}& \textbf{Shot}&\textbf{Score}&\textbf{Confidence RMSE}
\\[0.5ex] % inserts table
\hline
%heading
Open Variant  & ITP	& 2-way &-&0.393	& 0.746 \\
              & DFKI DKT& 2-way&-& 0.393 & 0.845\\
	           &ECNU & 2-way &-&0.464	& 0.736 \\
Closed Variant&IITP	& 2-way &-&0.286 & 0.807\\
	&IKM	& 2-way&-&0.536 &0.763\\
	&NileTMRG	& 2-way &-&0.536& 0.672\\
    \hline
	&\textbf{Baseline}	&- &-& \textbf{0.571}	 & - \\
   \arrayrulecolor{black}\hline
   
    & &&zero	&  \textbf{0.679}	& \textbf{0.513} \\
    & & 2-way&one	&\textbf{0.607}& \textbf{0.311} \\
     & &&few	&0.538 & \textbf{0.398} \\\cline{3-6}

	&&&zero	&\textbf{0.643}	&- \\
   &GPT-3.5-turbo&3-way&one &	\textbf{ 0.679} &- \\
   &&&few	&0.571&-  \\\cline{3-6}
	&&&zero	&0.429	&- \\
	&&2-step&one	&0.464&- \\
	&&&few	&0.500&-  \\\cline{3-6}
Openly close &&&zero	&\textbf{0.714*}& \textbf{0.178*} \\
      & & 2-way&one	& \textbf{0.643} & \textbf{0.220} \\
      & &&few	&\textbf{0.607}& \textbf{0.292} \\\cline{3-6}
		&&&zero	&0.393	& - \\
  	&GPT-4&3-way&one	&	0.357 &- \\
	&&&few	&0.357&- \\\cline{3-6}
	&&&zero	&0.357	& - \\
	&&2-step&one	&0.429	&- \\
	&&&few	&0.429	&- \\

\arrayrulecolor{black}\hline
\end{tabular} 
\vspace{-0.5cm}
\label{tab:subtask1_results}
\end{table}

The results for subtasks 1 and 2 are summarized in Tables \ref{tab:subtask1_results} and \ref{tab:subtask2_results} respectively. For rumour veracity prediction, GPT models introduce a substantial improvement as demonstrated in Table \ref{tab:subtask1_results}. 
Both GPT-3.5-turbo and GPT-4 outperform the baseline and all the 2017's participating teams for a number of classification schemes and shot settings. The corresponding values are shown in bold in the table.
The top best results (with an asterisk) for both the score and confidence RMSE are achieved by zero-shot GPT-4 using binary classification. They outperform the 2017's baseline by more than 25\% for accuracy score, and the 2017's best participating model "NileTMRG" by 33.2\% for score and 73.5\% for confidence RMSE. GPT-4 is already expected to be more refined and less erroneous compared to GPT-3.5-turbo.
The second best results introduce a gain increase of 18.9\% over the baseline, and are achieved by both (1) zero-shot GPT-3.5-turbo via binary classification, and (2) one-shot GPT-3.5-turbo via 3-way classification.
From our top 2 best results, it is confirmed that binary classification is the best classification method across all the experimented GPT variants, and using all the experimented metrics.
The third best score (with an 12.6\% increase over the baseline) is achieved by both (1) zero-shot GPT-3.5-turbo through 3-way classification, and (2) one-shot GPT-4 via binary classification.
It is worth mentioning that confidence RMSE numbers are only computed for binary classification to keep consistency with the 2017 scoring scripts.
Surprisingly, 2-step classification using prompting achieves the worst results among the old and new methods. The reason for such poor results is that most of the VLLM predictions were classified as \textit{"unverified"}, even for those examples that are \textit{"verified"} but \textit{"false"}. This suggests that it still difficult, even for VLLMs, to accurately classify a rumour as \textit{"verified"} or \textit{"unverified"}. This finding is even more flagrant for GPT-4 than GPT-3.5-turbo.
We also find that zero-shot outperforms the other counterparts. However, performance differences between zero-, one-, and few-shot are small and inconsistent across models and classification schemes. This finding was not encouraging to further explore one- and few-shot outputs for the other task.

For rumour stance classification, the results in Table \ref{tab:subtask2_results} are not as promising as those of veracity prediction. Both GPT-4 and GPT-3.5-turbo perform poorly using 3-way or 4-way classification methods, compared to the other teams. For 4-way classification specifically, none of the models was able to beat even the baseline. However, there is a slight hope for 3-way classification with GPT-4 where the model was able to outperform the baseline by 98\%, and ranked the third best among the best participating teams after "Turing" and "ECNU" models. This finding suggests that the effect of class imbalance is not only restricted to pre-trained models, but can also negatively impact generative models. VLLM stance classification results with \textit{"comment"} are worse than stance prediction outputs when the majority class is excluded.
\begin{table}[!t]
% \vspace{-1cm}
\begin{minipage}[t]{0.46\linewidth}\centering
\Small
\caption{Rumour SDQC Stance Classification Results: The first 8 rows are the participating teams in RumourEval 2017. The 2 baselines for 4-way and 3-way classification are on the 9th and 10th rows respectively.  Our best result, achieved by 3-way using GPT-4, could not beat the Turing team. Overall, 3-way classification by GPT-4 is the third best result.}
\vspace{-2em}
% \centering      % used for centering table 
\begin{tabular}{cc} 
% \hline          %inserts double horizontal lines
% \cline{2-9}
 \\ [0.3ex]
\hline
\textbf{Team}&\textbf{Score}
\\[0.5ex] % inserts table
\hline
DFKI DKT & 0.635 \\
ECNU	& 0.778 \\
IITP 	& 0.641 \\
IKM 	& 0.701 \\
Mama Edha &	0.749 \\
NileTMRG &	0.709 \\
\textbf{Turing} & \textbf{0.784} \\
UWaterloo & 0.780 \\
\hline
Baseline (4-way) &	0.741\\
Baseline (SDQ) & 0.391 \\
\hline
GPT-4 (4-way) & 0.580 \\
GPT-3.5-turbo (4-way)	& 0.620 \\
GPT-4 (SDQ)	 & 0.774 \\
GPT-3.5-turbo(SDQ)	& 0.650 \\
\arrayrulecolor{black}\hline
\end{tabular} 
\vspace{-0.05cm}
\label{tab:subtask2_results}
\vspace{-0.5cm}
\end{minipage}
\begin{minipage}[t]{0.53\linewidth}\centering
\begin{lstlisting} [label=lst12, language=HTML, caption={Subtask 2: Output Example for Multiclass Classifcation by GPT-3.5-turbo.} ]
'tweet_id': 524948206023880704,
'tweet_text': 'The latest @CBCNews on #Ottawa shootings. Gunman killed inside Parliament, soldier shot at War Memorial. http://t.co/kngapKTSCe #cdnpoli',
'tweet_class': 'true',
'reply_id': 524999801973506048,
'reply_text': "@Trial_Watcher1 @CBCAlerts @CBCNews And the  news I'm watching right now are saying so much different than what you're posting in here LOL",
'ground_truth': 'deny',
'prediction': ['deny', 'comment'],
'confidence': [0.8, 0.2],
'justification': "The reply includes phrases like 'so much different' and 'LOL', indicating disagreement and a mocking tone. This suggests a denial of the information in the source tweet and a comment on the discrepancy between the news being watched and the information in the tweet."
\end{lstlisting}
\vspace{-0.4cm}
\end{minipage}
\vspace{-0.5cm}
\end{table}
The example in \ref{lst12} illustrates a correct SDQC stance output generated by GPT-3.5-turbo using multiclass classification. The example shows that both the confidence scores and the justifications provide further explanations on why a particular decision is made, further manifesting the XAI abilities of VLLMs.
The original ground truth for the stance is \textit{"deny"}. Since multiclass is a novel extension to this task, no ground truth will match the generated prediction of \textit{"deny"} with 0.8 confidence score and \textit{"comment"} with 0.2 score. However, the provided justification can be very reasonable and convincing as shown in \ref{lst12} Line 9.
We think prompting-based VLLMs are promising solutions to the misinformation spread. Further experiments are still needed to unleash their full potential.

\vspace{-0.3cm}
\section{Discussion}
\label{discussion}
% Working with VLLMs for rumour verification solutions has its own complexities. Below is a discussion of the challenges, limitations, and potential future work.
% relevant to this research direction.

\vspace{-0.2cm}
\subsection{Challenges}
\label{challenges}
\hspace{-0.4cm}\textbf{Cost:} One limitation in our experiments is the cost. Although the models are not expensive and the cost is determined per token, OpenAI models are not entirely free. The company has different payment plans depending on the usage. However, if one is not careful, s/he can really spend a significant amount of money without even realizing. Our experiments are heavily relying on the trial and error approaches.\\
\textbf{Availability:}
The models are not always available to run. Despite being on the paid plan and because of traffic bottlenecks, the models kept hanging in the middle, causing us to stop and restarting the experiments, further spending more money.  We estimate that only 2/3 of our runs were successful on the first attempt. \\
% It is understandable that these models are hot topic, and many researchers are using them for their work causing traffic bottlenecks. We believe that these models will be better maintained with further technology advancement. \\
\textbf{Time Consumption:}
GPT-4 is slow, especially with large scale dataset. Imminent glitches are frequent, adding to the overall running time. The system can stop and restart by itself.
% , which will likely double the expected time to be consumed.
It took us 6 hours to generate predictions for 1000 examples. Although it is still relatively time-efficient compared to traditional training that can take weeks, generative AI should be a lot more flexible in terms of time consumption.\\
\textbf{Temperature Setting:}
Generative VLLMs are known for their randomness. While GPT models can be adjusted to guarantee result consistency (by setting the temperature to 0), this restricts VLLM capabilities of exploring newer approaches, hence improving its predictions. Choosing the optimized temperature poses a challenge because of the GPT consistency-creativity tradeoff. In our experiments, we performed a grid search of temperatures [0, 1] with 0.1 increments to determine the best value. However, the high temperature value suggests weak result consistency. A valued topic for future research is to find ways to tune the temperature while maintaining good prediction and tolerable consistency.
% =Version 2 =================================
% Generative VLLMs are known for their randomness. While GPT models can be adjusted to guarantee result consistency, this restricts VLLM capabilities
% of exploring newer approaches, hence improving its predictions. Tuning the temperature is not an easy task. A temperature of "0" will guarantee result consistency, but may have poor predictions.
% In our experiments, results started to improve when the temperature was increased (using a grid search of values between [0, 1] in increments of 0.1), but this
% induced  randomness and inconsistency as well. Future research is needed to find the best temperature setting to balance prediction and consistency.
\vspace{-0.3cm}
\subsection{Limitations and Future Work}
\vspace{-0.1cm}
\hspace{-0.4cm}\textbf{One Social Media Platform:}
Our experiments follow those of RumourEval 2017 in terms of dataset. The latter was collected from rumourous Twitter threads. It would be interesting to create a homogeneous unified dataset from well-known social media platforms like Facebook, Instagram, Twitter, Reddit, LinkedIn, Slack, Whatsapp, Discord and others. The unified dataset can be used as a benchmark for VLLM performance comparison.\\
\textbf{English Data:}
The dataset only accounts for English tweets, restricting it to one group of people. A potential extension to this work is to experiment on multilingual social media data, and examine the behaviour with prompt-engineering-based models.\\
% Will the accuracy scores of rumour veracity prediction change? By what factor? What about stance classification? These research questions need be exhaustively addressed.\\
% \textbf{Few-shot Prompting:}
% One of the main ideas we had originally planned is to conduct the very same experiments in a few-shot setting. Due to the cost and time constraints, we was not able to fully commit to this plan. However, few-shot prompting has a great chance of improving the results, especially those of rumour stance classification. VLLMs already have a massive amount of knowledge and great learning potentials. We believe leveraging few examples demonstrating support, deny, query and comment in the prompt could help the model better refocus on the desired output. A number of trial and experiments will be required to reach the optimal number of the required examples. Too many examples are as harmful as very few.\\
\textbf{Hallucinations and Biases:}
It is well-known fact that conversational VLLMs suffer from issues like hallucinations and biases. Throughout this work, we did not get the chance to heavily investigate and analyze the hallucinated justifications/confidence scores.
% that were generated in case of wrong predictions.
% Conversational models can be very convincing with their explanations.
To mitigate this phenomenon, measures need to be considered like additional clarity and specificity in the prompts, constraints and boundaries on the expected outputs and temperature tuning.\\
% We also believe few shot prompting can help include factual information within the prompt to ground its reasoning, hence preventing it from drifting into hallucinations.\\
\textbf{Other VLLMs:}
Only two GPT variants (GPT-4 and GPT-3.5-turbo) were experimented for rumour veracity and stance classification. An extension to this work is to explore other conversational VLLMs like Google's Bard/Gemimi/PaLM or Meta's LLaMA family. Advances in the generative AI field are fast, and new models emerge every few days. A performance comparison on a wider range of models can surely solidify this research line. \\
\textbf{Broadening scope:}
It would be useful to broaden our study of providing 
effective explanations. One starting point is
\cite{10.1007/978-3-030-88483-3_23}, which clarifies how specific language elements should matter.
We could deepen our analysis of stance, inspired by 
\cite{DBLP:journals/ci/SobhaniIZ19} which clarifies the importance of multiple instances of targeting within a social network. Studies with datasets regarding Covid-19 (a topic rife with misinformation) are also of interest; \citet{DBLP:journals/corr/abs-2105-13430} also discuss post-hoc vs. ante-hoc explanation, which may assist in further comparisons to misinformation work \cite{kim-yoon-2022-detecting}.
%%%%%%%%%%%%%%%%%%%%%%%%%%%%%%%%%%%%%%%%%%%%%%%%%%%%%%%%
\vspace{-0.3cm}
\section{Conclusion}
\vspace{-0.2cm}
\label{conclusion}
In an attempt to bridge the research gap between generative AI and the research efforts towards the limitation of misinformation spread, we propose to leverage prompt-engineering-based VLLMs, more specifically GPT-3.5-turbo and GPT-4, to extend the RumourEval task of SemEval 2017 workshop. We conduct 24 types of experiments for both subtasks: veracity prediction and rumour stance classification. For veracity prediction, three classification schemes are attempted per VLLM variant including binary, ternary and 2-step classifications. Zero-, one- and few-shot experiments are conducted. For stance classification, 4-way SDQC, 3-way SDQ and multiclass classifications were experimented per variant.  Our best results, for subtask 1, outperform all the precedent ones, including the baseline, for both accuracy and confidence RMSE. For subtask 2, the best results could not beat other models and ranked as third. All of our experiments generate confidence scores and natural language justifications for every prediction for trust and explainability purposes. 
% Our findings suggest that generative AI models are promising for the rumour evaluation domain, yet further investigation and research are sill required before being able to adopt them autonomously.
% To the end, we share  our implementation, experimental settings, results, analysis, challenges, limitations, and potential future work with the research community.
\vspace{-0.3cm}

\appendix

% \section{Example of math equation }
% %\label{appendix-customize-this-label}
% Binomial theorem: \cite{abramowitz1948handbook}
% \begin{equation}
% (x+y)^n=\sum_{\substack{k=0}}^{n}\dbinom{n}{k}x^{n-k}y^k
% \end{equation}

% All references should be stored in the file "references.bib".
% That call to use that file is in "cai.cls". 
% Please do not modify anything below this line.
\printbibliography[heading=subbibintoc]

\end{document}